\documentclass[conference]{IEEEtran}
\IEEEoverridecommandlockouts
\usepackage{cite}
\usepackage{amsmath,amssymb,amsfonts}
\usepackage{algorithmic}
\usepackage{graphicx}
\usepackage{textcomp}
\usepackage{threeparttable}
\usepackage{multirow}
\usepackage{enumitem}
\usepackage{booktabs}
\usepackage{subfigure}

\def\BibTeX{{\rm B\kern-.05em{\sc i\kern-.025em b}\kern-.08em
    T\kern-.1667em\lower.7ex\hbox{E}\kern-.125emX}}
\begin{document}

\title{Colorectal Polyp Detection in Real-world Scenario: Design and Experiment Study}

\author{\IEEEauthorblockN{Xinzi Sun$^{1}$, Dechun Wang$^{1}$, Chenxi Zhang$^{1}$, Pengfei Zhang$^{1}$, Zinan Xiong$^{1}$, Yu Cao$^{1}$, Benyuan Liu$^{1}$\\ 
Xiaowei Liu$^{2}$, Shuijiao Chen$^{2}$}
\IEEEauthorblockA{\textit{$^{1}$ University of Massachusetts Lowell, USA} \\
\textit{$^{2}$ Xiangya Hospital of Central South University, China}}}

\maketitle

\begin{abstract}
Colorectal polyps are abnormal tissues growing on the intima of the colon or rectum with a high risk of developing into colorectal cancer, the third leading cause of cancer death worldwide. Early detection and removal of colon polyps via colonoscopy have proved to be an effective approach to prevent colorectal cancer. Recently, various CNN-based computer-aided systems have been developed to help physicians detect polyps. However, these systems do not perform well in real-world colonoscopy operations due to the significant difference between images in a real colonoscopy and those in the public datasets. Unlike the well-chosen clear images with obvious polyps in the public datasets, images from a colonoscopy are often blurry and contain various artifacts such as fluid, debris, bubbles, reflection, specularity, contrast, saturation, and medical instruments, with a wide variety of polyps of different sizes, shapes, and textures. All these factors pose a significant challenge to effective polyp detection in a colonoscopy. To this end, we collect a private dataset that contains 7,313 images from 224 complete colonoscopy procedures. This dataset represents realistic operation scenarios and thus can be used to better train the models and evaluate a system's performance in practice. We propose an integrated system architecture to address the unique challenges for polyp detection. Extensive experiments results show that our system can effectively detect polyps in a colonoscopy with excellent performance in real time.
\end{abstract}


\section{Introduction}
Colorectal cancer (CRC) is the fourth most commonly diagnosed cancer and the third leading cause of cancer death worldwide in 2018 \cite{bray2018global}. In the United States, CRC is the third most common cancer, accounting for 9\% of all cancer incidence \cite{doi:10.3322/caac.21551}. Currently, colonoscopy is the primary and most effective method for screening for and preventing CRC. A colonoscopy is an outpatient procedure in which a tiny camera is navigated inside the large intestine (colon and rectum) to check for abnormalities and diseases. During the colonoscopy, abnormal growths, such as colorectal polyps which is an essential precursor of CRC, will be identified and removed. Therefore, polyp detection during colonoscopy plays a crucial role in CRC early detection and treatment.

\begin{figure}[htp]
\centering
\includegraphics[width=0.5\textwidth]{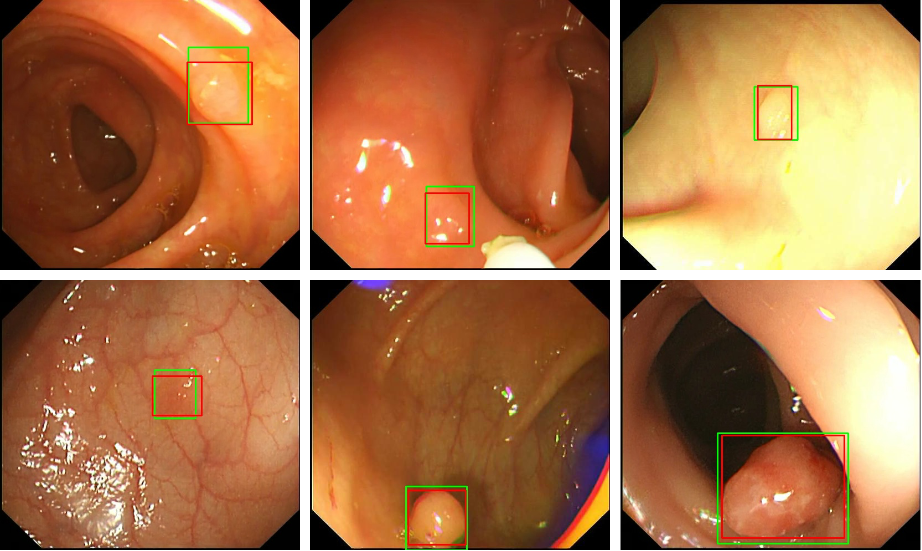}
\caption{Sample results of our polyp detection system. Green rectangles represent the prediction of the system and red rectangles represent the ground truth. Best viewed in color.} 
\label{fig:real application}
\end{figure}

A colon polyp is a clump of cells that grow on the lining of the colon. Polyps grow through rapidly dividing cells, similar to how cancer cells grow. This is why they can become cancerous, even though most polyps are benign. Some colon polyps can develop into CRC over time, which is often fatal when found in later stages. However, according to Leufkens et al.’s study \cite{leufkens2012factors}, 22\%-28\% of polyps and 20\%-24\% of adenomas are missed in colonoscopy due to various human factors. Therefore, there is a critical need for an efficient and accurate computer-aided colorectal polyp detection system that can assist physicians with localizing the polyps during a colonoscopy.

\begin{figure*}[htbp]
\centering
\includegraphics[width=\textwidth]{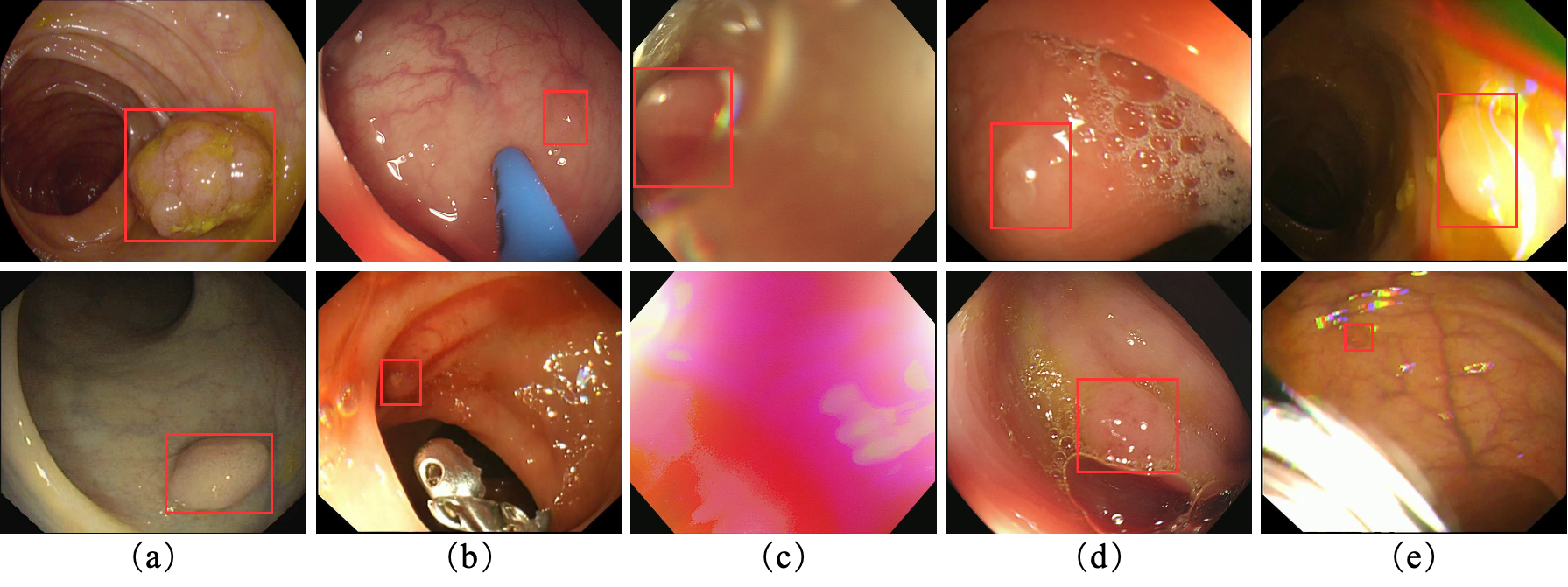}
\caption{Difference between public datasets and real colonoscopies. (a) images from public datasets; (b) to (e) are images from real colonoscopies; (b) images with medical instruments; (c) blurry images caused by different factors; (d) bubbles; (e) specularity.} 
\label{fig:polyps}
\end{figure*}

Previous polyp detection methods rely mainly on hand-crafted features such as color, texture or shape to extract specific patterns of polyps. However, due to the vast diversity in these features, traditional methods do not perform well for polyp detection. Recently, with the advances in deep learning, Convolutional Neural Network (CNN)-based approaches have been widely adopted for polyp detection and segmentation. For example, object detection methods such as Faster R-CNN \cite{ren2015faster} and YOLO \cite{redmon2016you} are used to find and indicate polyps with bounding boxes \cite{Mo2018AnEA,zhang2018polyp}. Semantic image segmentation methods such as U-Net \cite{ronneberger2015u}, Fully Convolutional Networks (FCN) \cite{long2015fully}. and SegNet \cite{badrinarayanan2017segnet} are employed to localize polys on the pixel level \cite{DBLP:journals/corr/abs-1806-01907,Li2017ColorectalPS,akbari2018polyp,Brandao2017FullyCN, wang2018development}.

While these CNN-based approaches have shown excellent performance in polyp detection and segmentation when trained and tested on publicly available datasets such as GIANA (Gastrointestinal Image ANAlysis) 2018 \cite{bernal2017comparative} and ETIS-Larib Polyp DB \cite{silva:hal-00843459}, there is a wide gap between images from real colonoscopy %operations 
and those in public datasets. Specifically, public datasets contain carefully chosen clear images with reasonably sized polyps that stand out from the background, without various artifacts that are often present in real operations. On the other hand, a large portion of the images in colonoscopy operations has different degrees of blurring due to the movement of the camera and intima, camera out-of-focus, or water flushes during an operation. Also, the images often contain various artifacts such as fluid, debris, bubbles, reflection, specularity, contrast, saturation, and medical instruments \cite{EAD2019endoscopyDatasetII,EAD2019endoscopyDatasetI}. Moreover, a wider variety of polyps with different sizes, shapes or textures can appear in a colonoscopy than those in the public datasets. Some representative images in the public datasets and real colonoscopy are shown in Fig. \ref{fig:polyps} to illustrate the differences. All these factors make polyp detection much more challenging in a real colonoscopy than on public datasets. In fact, we find that models optimally trained on public datasets suffer a significant performance degradation in real colonoscopy videos (F1-score drops from 95.77\% for the model trained on CVC-ClinicDB and 81.07\% for the model trained on ETIS-Larib Polyp DB to 70.45\% in real colonoscopy), frequently missing true polyps and falsely labeling artifacts (e.g., bubbles, specularity, instruments, undigested debris, color blobs in blurry images, etc) as polyps.

In this research, we collaborate with the Endoscopy Center of Xiangya Hospital of Central South University in China and collect a dataset that contains 7,313 images from 224 complete colonoscopy operations with pixel-level polyp and instrument annotations for each image. The large variation of polyps in terms of size, shape, and texture can substantially improve the sensitivity of the model trained on this dataset. In addition to the vast diversity of polyps, our dataset has a large fraction of images with various artifacts and blurry images caused by camera motion during the operations. Due to the presence of various artifacts in a colonoscopy, a single CNN-based polyp detector tends to generate a considerable number of false positives. While this problem may be alleviated by increasing the output threshold of the model, however, a threshold too high will inevitably lower the detection rate of true polyps. Meanwhile based on the finding that two of our recently developed CNN models have high consistency in localizing true polyps and low consistency on false detections, we propose an integrated polyp detection architecture that consists of a blurry detection module to filter out blurry images and an ensemble module to combine the results from an Anchor Free Polyp (AFP-Net) detector \cite{wang2019afp} and a U-Net with Dilation Convolution detector \cite{sun2019colorectal}. By removing blurry images from subsequent model inference, we can significantly reduce the processing time of the pipeline. By combining the results from two independent detectors, we can dramatically improve the specificity of our system without losing too much sensitivity. Fig. \ref{fig:real application} shows sample results of our integrated polyp detection system in real colonoscopy procedures.

In summary, the key contributions of this paper include: (i) We create a private dataset that contains 7,313 images from 224 complete colonoscopy procedures. Images in this dataset are highly representative of realistic operation scenarios compared to those well-chosen ones in the public datasets, and thus can be used to better train the models and evaluate the system's performance in practice. (ii) We propose an integrated system architecture that addresses the unique challenges for polyp detection in a colonoscopy. The system consists of a blurry detector that filters out blurry images to reduce processing time and two independent polyp detectors that are combined to enhance the accuracy. (iii) We train our models and test our proposed system on complete colonoscopy videos. Results show that our system can effectively detect polyps in colonoscopy operations with excellent performance in real-time at a rate of 23 frames per second.

\begin{figure*}[htbp]
\centering
\includegraphics[width=\textwidth]{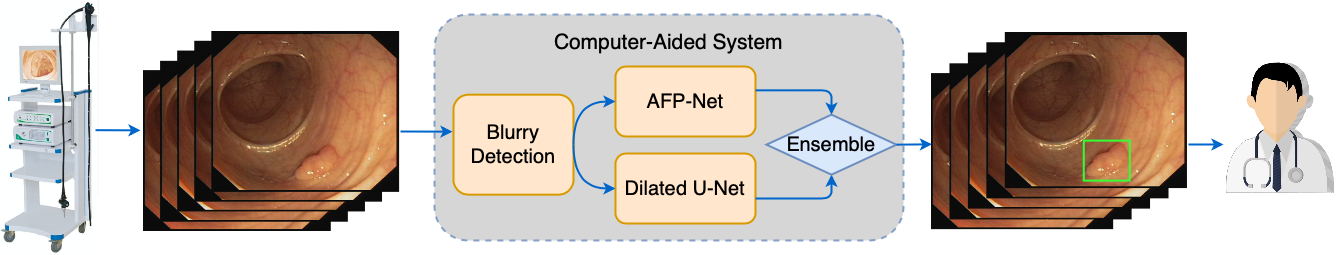}
\caption{The architecture of our polyp-detection system. It consists of a blurry detection module, two polyp detection modules, and an ensemble module.} 
\label{fig:system_architecture}
\end{figure*}

\section{Related Work}
Early polyp detection methods usually utilize hand-crafted features and a simple classifier to distinguish polyps from normal tissues. Gross \emph{et al.} \cite{Gross_polypsegmentation}, Bernal \emph{et al.} \cite{bernal2012towards,bernakL:wm-dova}, and Ameling \emph{et al.} \cite{ameling2009texture} used color, shape, or texture features as key factors to identify the locations of polyps. Ganz \emph{et al.} \cite{ganz2012automatic} and Mamonov \emph{et al.} \cite{mamonov2014automated} utilized shape or contours information for polyp segmentation. However, since most polyps and normal tissues have similar edge and color features, traditional hand-crafted-feature based methods can only detect polyps with several typical patterns, resulting in a poor performance in real applications. 

With the success of deep learning in computer vision, CNN-based polyp detectors have been investigated in the last few years. Mo \emph{et al.} \cite{Mo2018AnEA} applied a fine-tuned Faster-RCNN with VGG-16 as the backbone for polyp detection with a running time of 5 fps on a K40 GPU. In addition, Shin et. al \cite{shin2018automatic} proposed a post learning scheme to enhance the Faster R-CNN detector. This post-learning scheme automatically collects hard negative samples and retrains the network with selected polyp-like false positives, which functions in a similar way to boosting. Zhang \emph{et al.} proposed a two-step pipeline for polyp detection in colonoscopy videos. In the first step, they use a pretrained ResYOLO to detect suspicious polyps. The polyps are assumed to be stable between two consecutive frames. In the second step, a Discriminative Correlation Filter based tracking approach was proposed to leverage the temporal information. 

Polyp segmentation is another popular approach to localize polyps in colonoscopy videos on the pixel level. Inspired by U-Net, Mohammed \emph{et al.} \cite{mohammed2018net} proposed Y-Net which fuses two fully convolutional encoders followed by a fully convolutional decoder. By using two encoders with and without pre-trained weights, the model addresses the performance loss due to domain-shift from the pre-trained network (natural images) to testing (polyp data) and limited training data, which are two common challenges when employing supervised learning methods in medical image analysis tasks. Wang \emph{et al.} \cite{wang2018development} built an automatic polyp-detection system with three threads for polyp segmentation, with each thread performing the image segmentation model based on SegNet \cite{badrinarayanan2017segnet}, each thread needs less than 100 ms to process one frame, since the system can process up to 30 frames per second.

\section{System Design}

The architecture of our computer-aided polyp detection system is illustrated in Fig. \ref{fig:system_architecture}. It consists of a blurry detection module, two polyp detection modules, and an ensemble module. The system takes each single image frame in a colonoscopy as the input. The blurry detection module first filters out blurry frames caused by camera movement, out-of-focus or water flushes during a colonoscopy. The two polyp detection modules perform polyp detection simultaneously on the remaining frames and their results will be ensembled as the output shown on the colonoscopy image.

\subsection{Blurry Detection}

Our Blurry detection module is a CNN-based binary classifier. To train and test the model, we sample frames from videos of colonoscopy procedures and manually selected 400 images (200 blurry and 200 clear images) for training and 225 (100 blurry and 125 clear images) for testing. We use the pre-trained SquezzeNet \cite{iandola2016squeezenet} as the backbone and fine-tune it on the train set. Our model achieves 95.0\% of sensitivity and 96.0\% of specificity on the test set. Moreover, by testing the model on videos, we observe that our model can filter out most of the blurry frames. % The running time of our blurry detector is about 3ms on a single NVIDIA GeForce GTX 1080 Ti GPU.

%We sample frames from videos of real-world colonoscopy operation and manually selected 200 images for training and 100 for testing (1:1 for blur and clear). We used pre-trained SquezzeNet as backbone and fine-tuned a binary classifier on our own Blurry Colonoscopy image dataset. We get 90\% accuracy on blur test set and get 70\% on polyp dataset. The image in blur dataset is 100\% blur or clear, the image in polyp dataset contain partial blur. We also test the model on the video, our model can filter most of the extremely blurry frame and not sensitive to partial blurry. %The response time of our blurry detector is about 3ms.

\subsection{AFP-Net}

% Figure 2 illustrates our framework design of our first polyp detection model. It 
AFP-Net \cite{wang2019afp} is a fully convolutional network that classifies and localizes objects on each enhanced feature map. We use VGG16 \cite{simonyan2014very} as the backbone and select $k = 6$ feature maps from the backbone (conv4\_3, conv5\_3, conv6\_2, conv7\_2, conv8\_2, and conv9\_2) for detecting objects at different scales. To further increase the context information for small objects, we feed each feature map to a context module before forwarding them to the anchor-free detection heads. These detection heads are single-staged and have similar structures to the heads in SSD, where two parallel subnets are dedicated to classification and localization.

\subsection{U-Net with Dilated Convolution}

%The architecture of our second polyp detection model is shown in Figure 3. 
Inspired by U-Net \cite{ronneberger2015u}, we construct an end-to-end convolutional neural network that consists of a construction part (encoder) and an expansive part (decoder). The backbone we use in the encoder is Resnet-50 \cite{he2016deep}. We remove the down-sampling operation before the last stage of Resnet-50 to retain feature resolution and introduce dilated convolution in the last stage to enlarge the receptive field. 
\begin{figure}[htp]
\centering
\includegraphics[width=0.5\textwidth]{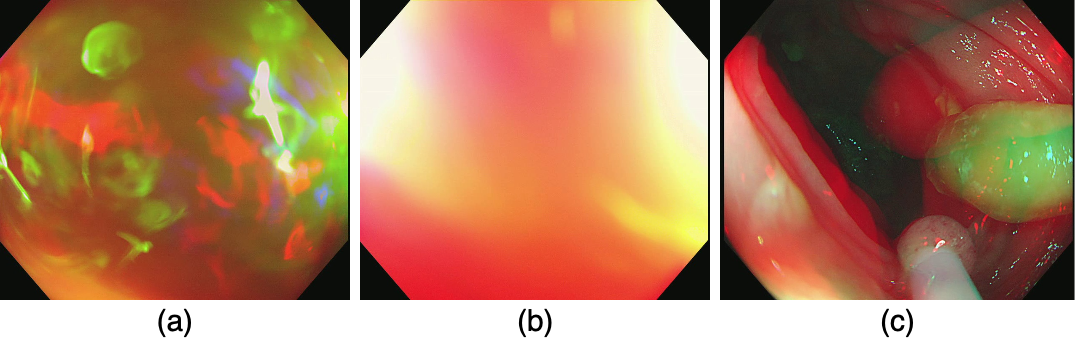}
\caption{Three causes of blurry imgaes. (a) water flush; (b) out-of-focus; (c) movement of the camera and intima} 
\label{fig:blurry_types}
\end{figure}
For the decoder, we use interpolation instead of transposed convolution to upsample each feature map back into the original size of the input, and combine these four feature maps to one block using concatenation and finally transform it into a 1-dimensional segmentation map. The segmentation map will be transformed into bounding-boxes with several post-processing procedures. The details of our blurry detection and two polyp detection modules can be found in \cite{wang2019afp,sun2019colorectal}.

% \subsection{Faster R-CNN}

% Faster R-CNN is the successor of fast R-CNN, it replaced the selective search algorithm with a separate region proposal network, which uses a sliding window over these feature maps, generate anchor boxes of different sizes and shapes, then generate region proposals, and made this step almost cost free. The predicted region proposals are reshaped using a RoI pooling layer, then used to classify the image within the proposed regions and predict the accurate position of the bounding boxes.

\subsection{Model Ensemble}

For every frame in a colonoscopy, the detection results from the two polyp detectors will be ensembled as follows. If the IoU (Intersection over Union) of the two resulting bounding-boxes is greater than 0.1, we keep one of the bounding-boxes while remove the other by applying non-maximum suppression. Otherwise, we ignore both of these results. By this ``AND" logic, as will be shown in Section 4, we are able to significantly reduce the false positive rate without trading off much sensitivity.

%Since both polyp detectors have high sensitivity on polyp detection and high consistency on polyp localization, this model ensembled strategy will not reduce the performance on recall by much. On the other hand, since these two polyp detectors have low consistency on false positive, the precision of our model will be considerably increased. In other words, false positives can be reduced without sacrificing true positives. 

\section{Dataset}

\subsection{Blurry Detection Dataset}
In order to train and test the blurry detector, we sample frames from videos of colonoscopy procedures and manually selected 400 images (200 blurry and 200 clear images) for training and 225 for testing (100 blurry and 125 clear images). The main causes of the blurry images in our dataset are the movement of the camera and intima, out-of-focus, and water flushes, which are shown in Fig \ref{fig:blurry_types}. Additionally, the blurry detection dataset contains both colonoscopy images with and without polyps.

\subsection{Public Dataset}

% \textbf{\textit{Comments: Use a table to summarize should be better and clearer}}

The public dataset consists of the datasets from GIANA (Gastrointestinal Image ANAlysis) 2018 \cite{bernal2017comparative} and ETIS-Larib Polyp DB \cite{silva:hal-00843459}. We split them into train and test set as follows:

\noindent{\bf Train set:} The train set consists of the following 3 datasets: 

\noindent({\romannumeral1}) 18 short videos from CVC-ClinicVideoDB (train set from GIANA Polyp detection Challenge) which contains 10,025 images of size 384 $\times$ 288. 

\noindent({\romannumeral2}) CVC-ColonDB: 300 images of size 574 $\times$ 500 from the train set of GIANA Polyp segmentation Challenge. 

\noindent({\romannumeral3}) 56 high definition images from GIANA Polyp segmentation Challenge with a resolution of 1,920 $\times$ 1080.

\noindent{\bf Test set:} The test set consists of the following 2 datasets:

\noindent ({\romannumeral1}) CVC-ClinicDB: 612 images of size 384 $\times$ 288 from the test set of GIANA Polyp Segmentation Challenge.

\noindent({\romannumeral2}) ETIS-Larib Polyp DB: 196 images of size 1,225 $\times$ 966.

\subsection{Private Dataset}

To strengthen the robustness of our model in a real-world clinical setting, we collected 224 videos of complete colonoscopy procedures at the Endoscopy Center of Xiangya Hospital of Central South University from July to November 2019. These videos are split into the train set and test set. We extract 96 and 239 short video clips that contain polyps from the training and testing videos respectively. These videos contain a large variety of polyps in terms of shape, size, color, texture as well as visual angles. To build the private dataset at the image frame level, we manually annotated 4,825 images with 4,535 polyps and 2,344 instruments from the training videos to form the train set, and 2,488 images with 2,688 polyps and 593 instruments from the testing videos to form the test set. In total, our private dataset contains 7,313 images, including 7,223 polyps and 2,937 instruments. These images and annotations were verified by experienced gastroenterologist from Xiangya Hospital of Central South University.

% To strengthen the robustness of our model in a real-world clinical setting, we collected 224 videos of complete colonoscopy procedures including the insertion and withdrawal phases from a hospital (name omitted for double-blind review) %the Endoscopy Center of Xiangya Hospital
%from July to November 2019. From these 224 videos, we extract 335 short video clips that contain polys. These videos contain a large variety of polyps in terms of shape, size, color, texture as well as visual angles. At the frame level, we manually annotated 7,313 images, including 7,223 polyps and 2,937 instruments. We split these images into the train set and test sets. To better examine the generalization performance, images in the train and test sets are chosen from different colonoscopy procedures. The resulting train and test sets contain 4,825 images with 4,535 polyps and 2,488 images with 2,688 polyps, respectively.

Compared to the public datasets, our private dataset is highly representative of real colonoscopy operations, containing more artifacts, and medical instruments. More importantly, some polyps are annotated in relatively blurry images caused by camera motion, out-of-focus, or water flushing in a colonoscopy. Additionally, all these 224 videos of complete colonoscopy were obtained from different types of colonoscopy equipment, such as Olympus CV-290 and Olympus CSV-290SL. Therefore, those images may vary in color cast, chromatic aberration, and resolution. All these factors contribute to a more challenging and realistic scenario for polyp detection. For this reason, we believe that models trained on our dataset will perform better in a real colonoscopy, and the performance evaluated on our private dataset will be a better measurement of the system's effectiveness.

%But, since the private dataset represent the real-world clinical environment, the performance evaluated on the private dataset is more convincing than on the public dataset.

% \subsection{Data Augmentation}
% Data augmentation is an important technique that has been widely used in machine learning pipeline. By introducing variations of images, such as different orientation, location, scale, brightness, etc, to existing data, we can increase the robustness and reduce over-fitting of our model. We apply some basic augmentation methods such as random rotation, random horizontal, vertical flip, and random zoom. We also apply tilt (skew), shearing, random distortion, since the tortuous appearance of the intima caused by the movement of colonoscopy lens and the colon. Besides, we apply random contrast and random brightness change to simulate different lighting environments or different photography equipment which may occur during colonoscopy procedures.

\begin{table}[tp]
\centering
\caption{Comparison between blurry detectors with different backbone.}
\fontsize{10}{12}\selectfont
\begin{tabular}{cccc}
\toprule
Backbone& Sensitivity& Specificity& fps\\
\midrule
Resnet-50 \cite{he2016deep} & 95.0\% & 98.4\%& 160\\
VGG16 \cite{simonyan2014very} &  90.0\% & 98.4\%& 490\\
SqueezeNet \cite{iandola2016squeezenet} & 95.0\% &96.0\% & 310\\
\bottomrule
\end{tabular}
\label{tab:backbone comparison}
\end{table}

\section{Experiments and Results}

\renewcommand{\arraystretch}{1} %控制行高
\begin{table*}[tp]
  \caption{Comparison between our system and previous methods on public datasets.}
  \centering
  \fontsize{10}{12}\selectfont
  \begin{threeparttable}
  \label{tab:public_performance_comparison}
    \begin{tabular}{ccccccc}
    \toprule[1.5pt]
    \multirow{2}{*}{Methods}&
    \multicolumn{3}{c}{ CVC-ClinicDB}&\multicolumn{3}{c}{ ETIS-Larib Polyp DB}\cr
    \cmidrule(lr){2-4} \cmidrule(lr){5-7}
    &Precision &Recall &F1-score &Precision &Recall &F1-score\cr
    \midrule
    % SNU \cite{bernal2017comparative} &26.80&26.40&26.50&10.20&9.60&9.70\cr
    % PLS \cite{Riegle:how} &28.70&76.10&41.60&15.80&57.20&24.90\cr
    CVC-Clinic \cite{bernakL:wm-dova} &83.50&83.10&83.30&10.00&49.00&16.50\cr
    ASU \cite{tajbakhsh2015automated} &97.20&85.20&90.80&-&-&-\cr
    OUS \cite{bernal2017comparative} &90.40&94.40&92.30&69.70&63.00&66.10\cr
    CUMED \cite{Chen:2016:DCN:3015812.3015985} &91.70&98.70&95.00&72.30&69.20&70.70\cr
    Faster R-CNN \cite{Mo2018AnEA} &86.60&98.50&92.20&-&-&-\cr
    FCN \cite{Li2017ColorectalPS} &89.99&77.32&83.01&-&-&-\cr
    FCN-8S \cite{akbari2018polyp} &91.80&97.10&94.38&-&-&-\cr
    FCN-VGG \cite{Brandao2017FullyCN} &-&-&-&73.61&86.31&79.46\cr
    
    Dilated U-Net\cite{sun2019colorectal} &96.71&95.51&96.11&80.48&81.25&80.86\cr
    AFP-Net \cite{wang2019afp} &99.36&96.44&97.88&88.89&80.77&84.63\cr
    \midrule
    Our system &98.85  &92.88 &95.77 &91.02 &73.08 &81.07 \cr
    \bottomrule[1.5pt]
    \end{tabular}
    \end{threeparttable}
\end{table*}

\renewcommand{\arraystretch}{1} %控制行高
\begin{table*}[tp]
  \caption{Comparison of our models on the private dataset.}
  \centering
  \fontsize{10}{12}\selectfont
  \begin{threeparttable}
  \label{tab:private_performance_comparison}
    \begin{tabular}{cccccccc}
    \toprule[1.5pt]
    \multirow{2}{*}{Methods}&
    \multicolumn{7}{c}{ Private Dataset}\cr
    \cmidrule(lr){2-8}
    &TP &FP &FN &Precision &Recall &F1-score &F2-score\cr
    \midrule
    Dilated U-Net (public)& 1746 & 752 & 927  &69.90  &65.32 &67.53 &66.19\cr
    Dilated U-Net& 1949& 617& 725& 75.95& 72.88& 74.38& 73.47\cr
    \midrule
    AFP-Net (public) & 1718 & 418 & 956 & 80.43  & 64.25 & 71.43 & 66.94\cr
    AFP-Net& 2117 & 319 & 557 &86.90  & 79.17 & 82.86 & 80.60\cr
    \midrule
    Our system (public) & 1572 & 217 & 1102  &87.87  &58.79 &70.45 &62.96\cr
    Our system  & 1885& 132 & 789 &93.46  &70.49 &80.37 &74.14\cr
    \bottomrule[1.5pt]
    \end{tabular}
    \end{threeparttable}
\end{table*}

\subsection{Blurry Detection Backbone Comparison}

We have experimented with building the blurry detection model with different backbones. Table \ref{tab:backbone comparison} shows the results of different backbones for the sensitivity, specificity, and running time. From Table \ref{tab:backbone comparison}, we can see that SqueezeNet \cite{iandola2016squeezenet} can achieve more than 95\% on both sensitivity and specificity with a per-frame processing time of only 3ms, representing the best trade-off between accuracy and time consumption. 

\subsection{Results on Public Datasets}

Table \ref{tab:public_performance_comparison} compares our results with current state-of-the-art methods on CVC-ClinicDB and ETIS-Larib Polyp DB. In this part, the AFP-Net, Dilated U-Net, and our integrated system are trained only on the public train set. Experiment results show that both polyp detectors and our integrated system significantly outperforms previous methods in precision, recall, and F1-score.

\subsection{Results on Private Dataset}

\begin{figure*} \centering    
\subfigure[] {
 \label{fig3:a}     
\includegraphics[width=0.95\columnwidth]{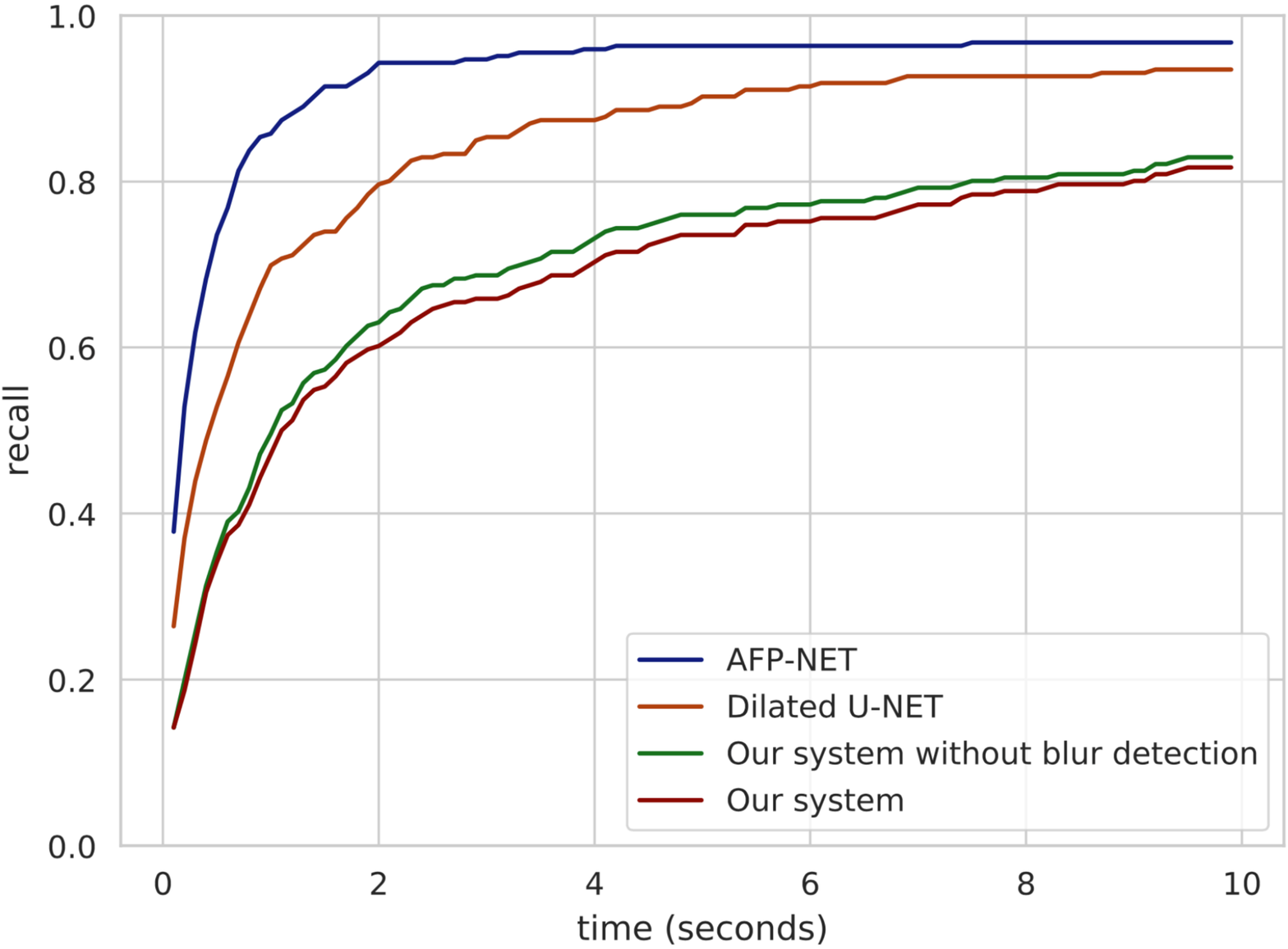}  
}     
\subfigure[] { 
\label{fig3:b}     
\includegraphics[width=0.95\columnwidth]{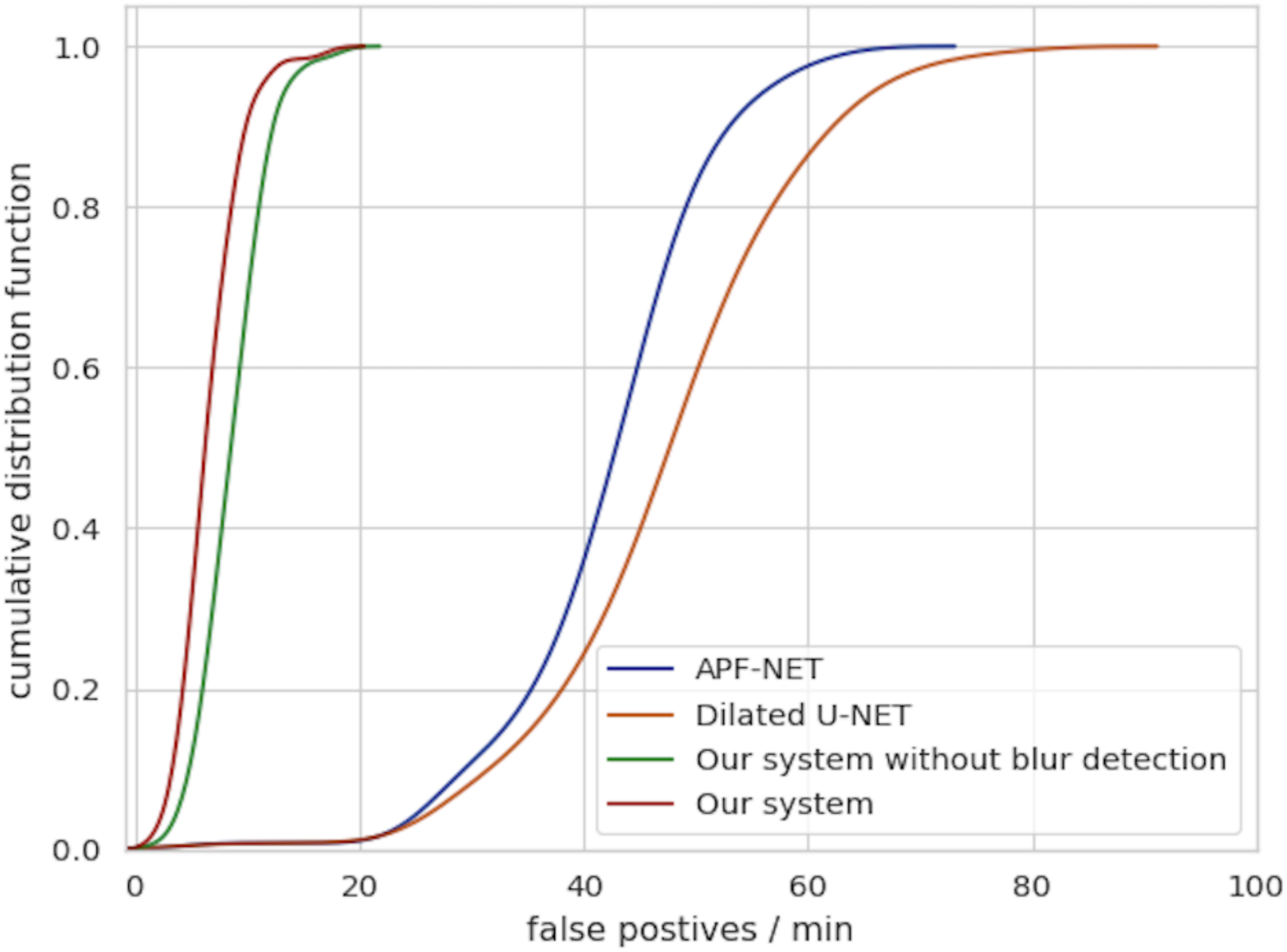}     
}    
\caption{Comparison between models on videos of colonoscopy. (a) recall as a function of time for different models to find a polyp. (b) cumulative distribution function (CDF) of the number of false positives generated per minute in each video for different models. Best viewed in color.}     
\label{fig3}     
\end{figure*}

In this part, we test our system at both image and video levels. All models are trained on the dataset that is a combination of the public and private train sets.

\noindent \textbf{Results on images:}
Table \ref{tab:private_performance_comparison} shows the results of our models on the private test set. %(still images)
Models labeled as ``(public)'' are trained only on the public datasets. From Table \ref{tab:private_performance_comparison}, we can observe that when trained on the combination of public and private datasets, the models significantly outperform their counterparts trained only on the public datasets. This is mainly due to the limited variety of polyps existing in the public datasets. As a result, models trained on these public datasets are not capable of detecting polyps of vast different varieties that may appear in a colonoscopy. This result clearly demonstrates the need and importance of a dataset that is representative of real colonoscopy operations.
% we can see that our private dataset can enhance the performance of our model significantly in real applications.

% \subsubsection{Results on video clips with polyps}
\noindent \textbf{Results of recall on videos:}
%For the computer-aided polyp detection system, one of the most important tasks in real-world colonoscopy operations is to detect polyps before physicians find them.
In real clinical practice, it is desirable to not only detect all polyps that appear in an operation but also to find each polyp as quickly as possible.
%perform the task in a real time fashion. 
Therefore, we design an experiment to measure how quickly our system can detect a polyp after its first appearance. All the experiments are carried out on 239 short video clips with polyps from testing videos.
%All the experiments are carried out on 239 short video clips with polyps. 
The results of each polyp detector and the integrated system are shown in Fig. \ref{fig3:a}. We can observe that our system can detect more than 60\% of the polyps within the first two seconds after a polyp appears and more than 80\% in ten seconds.

\noindent \textbf{Results of precision on videos:}
One of the most difficult challenges of applying a computer-aided system in real-world clinical practice is that models may generate a large number of false positives. In this part, we use video clips without polyps from testing videos of complete colonoscopy procedures to measure how many false positives our system generates. Since false positives within a short time span are usually highly correlated, for those false positives falling within 6 frames (100ms), we only count them as one incident. Fig. \ref{fig3:b} shows the cumulative  distributions function of the number of false positives generated per minute in each video for different models. From this figure, we can observe that we can eliminate most of the false positives from more than 40 per minute for the two individual detectors to less than 6 per minute for the integrated system. The number of false positives is smaller with blurry detection than without, as these false positives cause by artifacts on a blurry images (e.g. color blobs) are eliminated.

\subsection{Running Time}

The inference time of our blurry detector, AFP-Net, and Dilated U-Net are around 3ms, 20ms, and 20ms, respectively. Therefore, the running time is around 3ms for a blurry frame and around 43ms (23fps) for a clear frame. All the tests are performed on a single NVIDIA GeForce GTX 2080 Ti GPU and Intel i7-9700K CPU with PyTorch 1.4.0.
%The running time of our system for one frame is around 40ms (23fps). All the tests are performed on a single NVIDIA GeForce GTX 2080 Ti GPU.
% \subsection{Ensemble model's results on private dataset (with and without pre-processing)}

% \begin{figure}
% \centering
% \includegraphics[scale=0.19]{fp.pdf}
% \caption{Comparison between models on videos of colonoscopy procedures. (a) recall as a function of time for different models to find a polyp. (b) distributions of the number of false positives generated per minute in each video for different models.} \label{fig:fp}
% \end{figure}

\subsection{Discussion}

From Table \ref{tab:public_performance_comparison} and Table \ref{tab:private_performance_comparison}, we can observe that AFP-Net and Dilated U-Net, while well trained on the public datasets, suffer a significant performance degradation on our private dataset. The F1-score of AFP-Net drops from 97.88\% on CVC-ClinicDB and 84.63\% on ETIS-Larib Polyp DB to 71.43\% on our private dataset. Similarly, the F1-score of Dilated U-Net declines from 96.11\% on CVC-ClinicDB and 80.86\% on ETIS-Larib Polyp DB to 67.53\% on the private dataset. These results demonstrate the impact of the wide gap between the colonoscopy images from real medical operations and those in the public datasets on the model performance. After we re-train these two polyp detectors on our private train set, the F1-score of AFP-Net and Dilated U-Net improve significantly by 6.85\% and 11.43\%, respectively. Meanwhile, the F1-score and F2-score of our integrated system improves by 9.92\% and 11.18\%. These results clearly show that images in our private dataset represent realistic operation scenarios compared to those in public datasets, and thus can help us better train the models and improve the system’s performance in practice.

Additionally, from Table \ref{tab:private_performance_comparison}, we can see that, with a moderate decrease in recall, the precision of our system improves notably from 75.95\% for Dilated U-Net and 86.90\% for AFP-Net to 93.46\%. This result shows that the ensemble of two models can effectively reduce the false positives in the detection. The reason to trade off recall for better precision is that we do not need to detect a polyp in every frame in real clinical practice. Instead, a polyp only needs to be detected in one of the frames to warn the physician during an operation. From Fig. \ref{fig3:a}, each of our polyp detectors finds more than 90\% of polyps when running independently. However, based on the CDF plots shown in Fig. \ref{fig3:b}, they also produce a large number of false positives (approximately 40 to 60 per minute). After the ensemble module, the false positives rate drops significantly from 40 per minate to around 8 per minute. Also, with the blurry detection module, the number of false positives further declines to around 5 per minute. Therefore, the blurry detector and ensemble module can effectively reduce the false positive rate without losing much sensitivity for polyps detection in real colonoscopy procedures.

% \subsubsection{Image-level results}
% From the results in Table \ref{tab:public_performance_comparison}, we find that our model achieves a outstanding performance on public datasets if we only train model on it. However, from the results in Table \ref{tab:private_performance_comparison}, we can see that the performance drops significantly in real-world clinic circumstance. After train our model on the private dataset, the performance of our system is enhanced markedly, which shows that our private dataset can help the computer-aid system get better performance in real application.

% \subsubsection{Video-level results}

% From Figure \ref{fig3:a}, we can find that each of our polyp detectors could detect more than 90\% of polyps when running independently. However, they will also produce approximately 40 false positives per minute. After we applying the blurry detection module and ensemble the results of each polyp detectors, the number of false positives drops to around 5 per minute. This results indicates that our blurry detection module and ensemble strategy can reduce the most of the false positives without decimating the performance on recall in real application.

\subsection{Ensemble According to Polyp Size}
\begin{figure}[htp]
\centering
\includegraphics[width=0.48\textwidth]{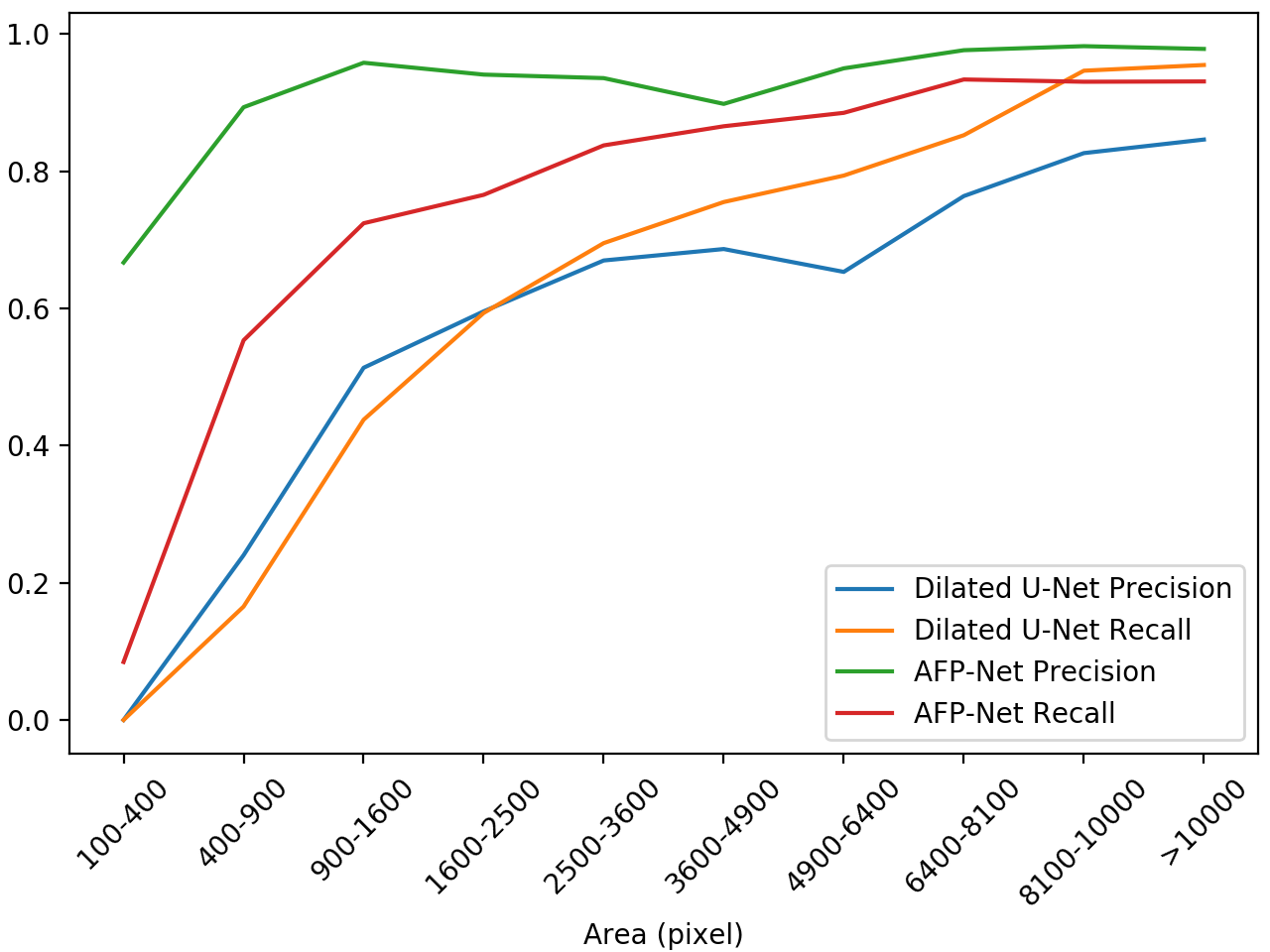}
\caption{Recall and precision as a function of the size of the predicted bounding-box for different models to find a polyp. All the experiments are processed on the private test dataset. Best viewed in color.} 
\label{fig:pre_recall_area}
\end{figure}

In real applications, we observe that different models are sensitive to the size of polyps. From Fig. \ref{fig:pre_recall_area}, we find that AFP-Net achieves a much better result than dilated U-Net on both precision and recall when polyp size is small. Therefore, in this section, we further introduce a more effective ensemble method with better performance and design a series of experiments to demonstrate its efficacy.  

\begin{figure}[htp]
\centering
\includegraphics[width=0.5\textwidth]{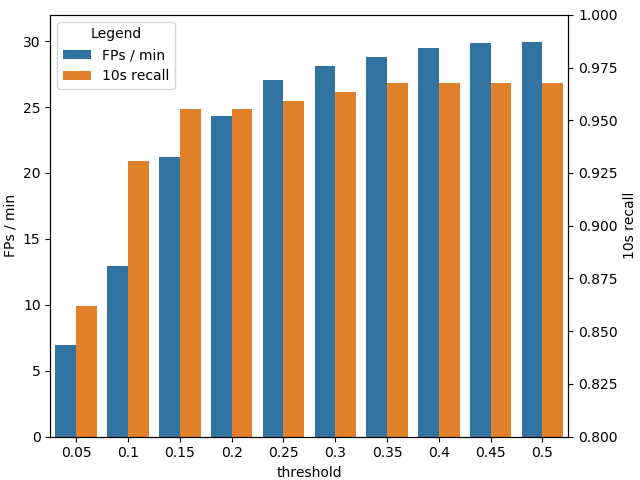}
\caption{Performance of our system with the new ensemble method on test videos. Blue bars represent the recall of our system with different thresholds of edge length. Orange bars represent the number of false positives generated by our system per minute with different thresholds of edge length. Best viewed in color.} 
\label{fig:new_ensemble_result}
\end{figure}

Based on the above observation, we redesign the ensemble module by introducing the length of the short edge of the prediction bounding-box as an extra parameter. When the short edge is less than a particular threshold, we only retain the outputs of AFP-Net, based on the fact that AFP-Net performs better than dilated U-Net when polyp size is small, while the performance gap narrows as the polyp size increases. To find the best threshold, we carry out a series of experiments to test the performance of the improved ensemble method on different thresholds. Here, we use the ratio of the length of the short edge of the prediction bounding-box and that of the input image as the parameter. The experiment results are shown in Fig. \ref{fig:new_ensemble_result}. By introducing this new parameter, physicians can adjust the sensitivity of the system to detect a polyp, especially for small polyps. For example, we can observe that by setting the threshold to 0.1, the recall of our polyp detection system can be dramatically improved from 80\% to 94\% with 13 false positives per minute. Additionally, the running time for processing a single frame is reduced to 23ms (time consumption the AFP-Net) when the size of the polyp is smaller than the configured threshold.

\section{Conclusion}

In this research, we collect a private dataset that contains 7,313 images from 224 complete colonoscopy operations with pixel-level polyp and instrument annotations. Images in our dataset represent realistic operation scenarios and thus can be used to better train a model and evaluate its performance. Additionally, we propose an integrated system that consists of a blurry detector and two polyp detectors. The results of two polyp detectors are ensembled to enhance the polyp detection performance. Results show that our system achieves excellent performance and can effectively detect polyps in real colonoscopy operations with a high accuracy in a real-time fashion.

%In this paper, we propose an integrated system for colorectal polyp detection which consists of a blurry detector and two polyp detectors. The results of two polyp detectors are ensembled to enhance the performance of our system. Additionally, we collected a private dataset which covers more kinds of polyp and circumstance appearing in the real-world clinical environment. Furthermore, we did extensive experiments on public datasets and our private dataset to show the performance of our system in real-world colonoscopy procedures. Our polyp detection system achieves great performance on both the public dataset and our private dataset.

\bibliographystyle{IEEEtran}
\bibliography{ref}

\end{document}